\documentclass[letterpaper, 10 pt, conference]{ieeeconf}  %

\IEEEoverridecommandlockouts                              %

\usepackage{graphicx} %
\usepackage{times} %
\usepackage{amsmath} %
\usepackage{amssymb}  %
\usepackage{xcolor}
\usepackage{multicol}
\usepackage{subfig}
\usepackage{multirow}
\usepackage{makecell}

{

\makeatletter
\let\NAT@parse\undefined
\makeatother
\usepackage{hyperref}

\newcommand{\vecXX}[1]{{\mathbf {#1}}}
\def \vecp {{\vecXX{p}}}
\def \vecq {{\vecXX{q}}}
\def \vecv {{\vecXX{v}}}
\newcommand{\argmax}{\operatornamewithlimits{arg\ max}}

\newcommand{\figref}[1]{Figure~\ref{#1}}
\newcommand{\tabref}[1]{Table~\ref{#1}}
\newcommand{\secref}[1]{$\S$\ref{#1}}

\title{\LARGE \bf
SafePicking: Learning Safe Object Extraction via Object-Level Mapping
}

\author{Kentaro Wada, Stephen James, Andrew J. Davison%
\\
Dyson Robotics Laboratory, Imperial College London%
\\
{\tt\small \{k.wada18, slj12, a.davison\}@imperial.ac.uk}%
}

\begin{document}

\maketitle
\thispagestyle{empty}
\pagestyle{empty}

\begin{abstract}
Robots need object-level scene understanding to manipulate objects while reasoning about contact, support, and occlusion among objects. Given a pile of objects, object recognition and reconstruction can identify the boundary of object instances, giving important cues as to how the objects form and support the pile. In this work, we present a system, \textbf{\textit{SafePicking}}, that integrates object-level mapping and learning-based motion planning to generate a motion that safely extracts occluded target objects from a pile. Planning is done by learning a deep Q-network that receives observations of predicted poses and a depth-based heightmap to output a motion trajectory, trained to maximize a safety metric reward. Our results show that the observation fusion of poses and depth-sensing gives both better performance and robustness to the model. We evaluate our methods using the YCB objects in both simulation and the real world, achieving safe object extraction from piles.
\end{abstract}

\section{INTRODUCTION}

6D manipulation of objects enables robots to alter object poses to a target state, to achieve tasks such as part assembly~\cite{Stevvsic:etal:RAL2020, Zakka:etal:ICRA2020}, object extraction from clutter~\cite{Wada:etal:CVPR2020, Zeng:etal:ICRA2017}, and arrangement in a specific configuration~\cite{Gao:Tedrake:ARXIV2019, Manuelli:etal:ISRR2019}. Traditionally, manipulation pipelines have been composed of perception that builds explicit scene representation with objects (e.g., 6D poses) followed by planning that searches a collision-free arm trajectory using the representation. As an alternative, a learning-based approach has emerged recently, which directly infers actions from observations (usually raw sensor information) with implicit scene understanding.

Although the traditional pipeline has been successful in structured environments, robotic manipulation in cluttered environments is still challenging due to close contacts and occlusions among objects. Prior work tackled this by combining state-of-the-art object pose estimation and collision-based motion planning~\cite{Wada:etal:CVPR2020, Zeng:etal:ICRA2017};  however, this pipeline has struggled to safely and efficiently extract occluded objects in a single step, causing potentially destructive effects on the objects, which is particularly important for fragile objects.

\begin{figure}[t]
  \centering
  \includegraphics[width=\linewidth]{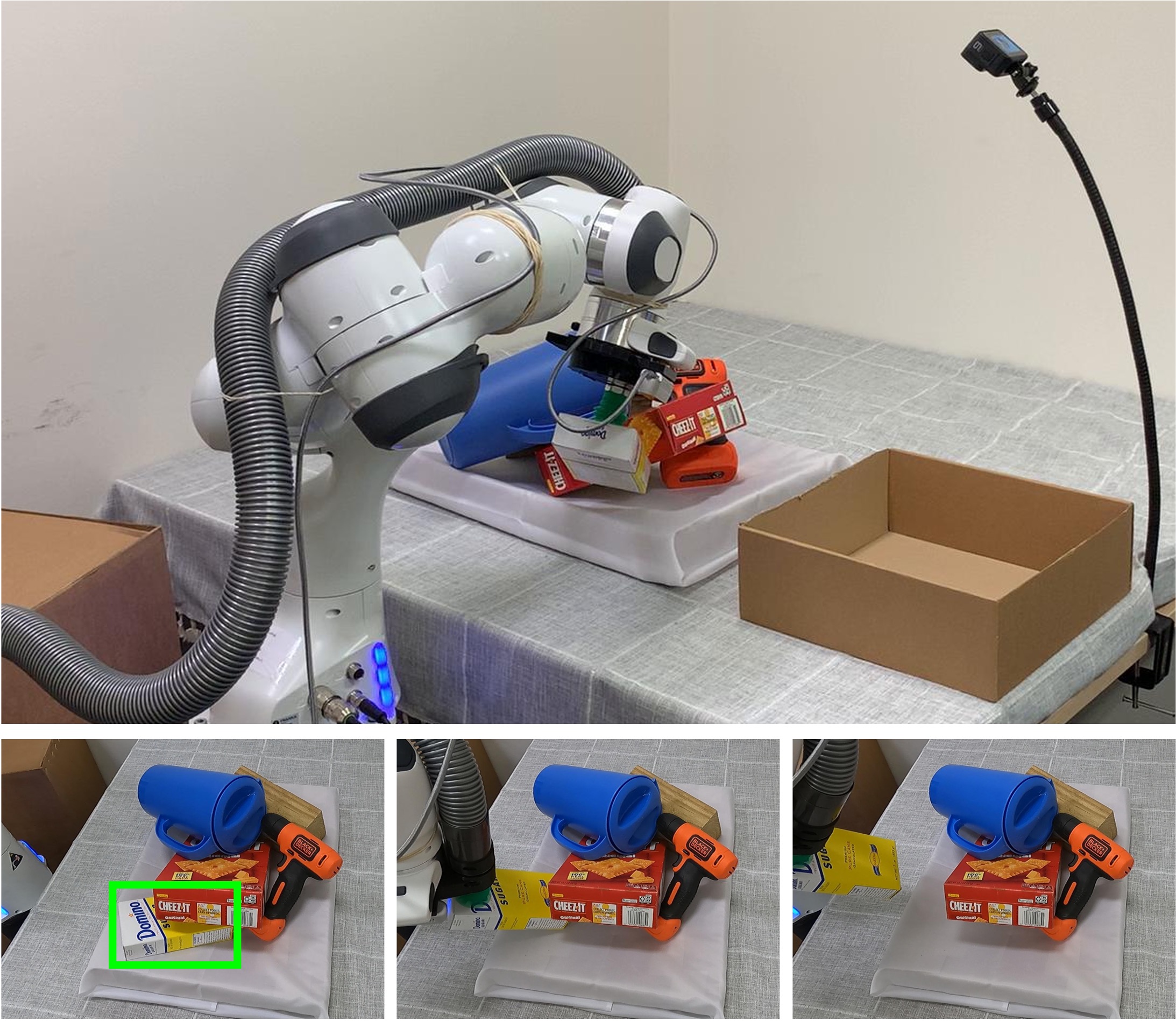}
  \caption{\textbf{SafePicking}\small{ extracts target objects with minimum destruction by generating a safe end-effector trajectory given a raw depth observation and object poses from an object-level map.}}
  \label{figure:teaser}
  \vspace{1mm}\hrule\vspace{-2mm}
\end{figure}

For this challenge in the traditional pipeline, we replace the collision-based motion planning with learning-based planning~\cite{Devin:etal:CORL2018, Kalashnikov:etal:CORL2018, Levine:etal:IJRR2018}, which receives as input some observation (e.g., images, object poses) and predicts the next best action to take. To avoid undesirable consequences on objects in a pile, we train the model with reinforcement learning while penalizing the translations of non-target objects during the extraction of a target object (e.g., \figref{figure:teaser}).

Despite the use of raw sensor observation being common in prior work, semantic information such as object poses can give important cues as to how robots should manipulate objects.  Object extraction requires a proper understanding of the occluded parts of objects, and we empirically observe that a learned model given object poses as input generates better motions even with errors in the pose estimate.

On the other hand, however, when the pose estimate has errors such as misdetection and pose-difference, a model that only uses object poses (as in the learning-based game agents~\cite{Baker:etal:ICLR2020, Brockman:etal:ARXIV2016}) would only deteriorate its performance. To handle these errors, we introduce raw observation along with object poses, which enables a model to gain high performance from the pose information as well as robustness from the raw observation.

\begin{figure*}[t]
  \centering
  \includegraphics[width=\linewidth]{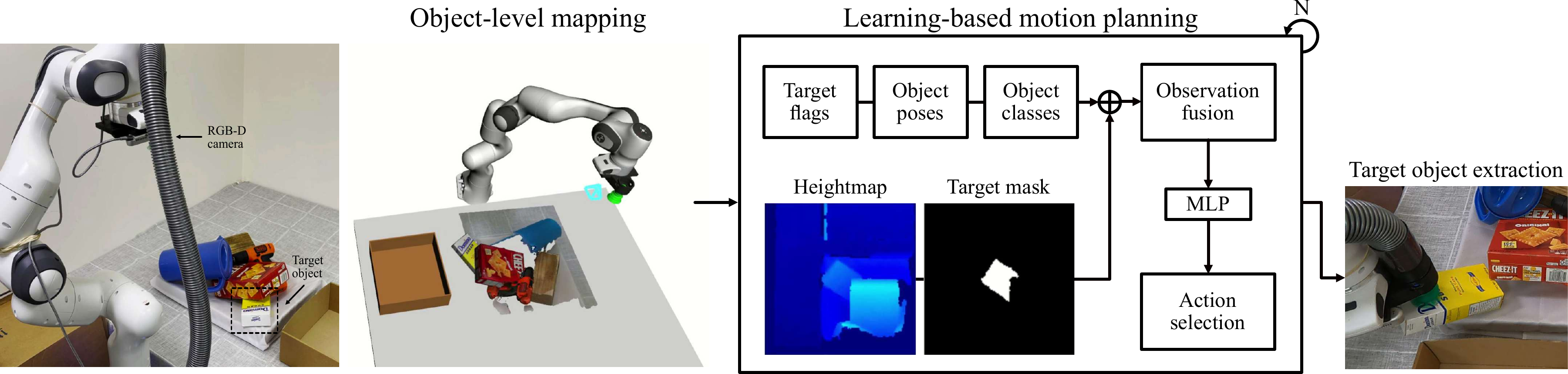}
  \caption{\textbf{System overview}, \small{which consists of 1) object-level mapping with volumetric reconstruction and pose estimation of detected objects with an on-board RGB-D camera, and 2) learning-based motion planning with a model that recursively generates end-effector relative poses with N steps for object extraction using estimated object poses and a raw depth observation in the form of a heightmap.}}
  \label{figure:overview}
  \vspace{1mm}\hrule\vspace{-3mm}
\end{figure*}

Our system, \textbf{\textit{SafePicking}}, shown in \figref{figure:overview} is composed of 1) \textbf{object-level mapping} to build a map with object poses while exploring a target object, and 2) \textbf{learning-based motion planning} to generate an end-effector 6D trajectory from raw (depth images) and pose observations (predicted object poses). The grasp point is determined to be the centroid of the visible surface of a target object, and the observations are transformed into the grasp point coordinate to be agnostic to the grasp point (canonicalization).  This combination of object-level mapping and canonicalization makes the subsequent learned model general to the object position in a workspace, enabling the model to learn faster.

To our best knowledge, this is the first work that tackles safe object extraction, in which a robot picks occluded target objects with a single grasp minimizing the destructive effect to the surrounding objects. In the experiment, we demonstrate our integrated system in the real world.

In summary, the main contributions of this paper are:
\begin{enumerate}
  \item \textbf{Introducing safe object extraction} as a novel manipulation task, where a robot extracts a target object with a single grasp while minimizing destructive effects.
  \item \textbf{Fusion of a raw observation and object poses} in learning-based motion planning, achieving high performance and robustness to estimation errors.
  \item \textbf{Integrated robotic manipulation system} that demonstrates safe object extraction in the real world.
\end{enumerate}

\section{Related Work}

\subsection{Manipulation with visual scene understanding}

Manipulation of a grasped object has been a long-standing problem that dates back to the beginning of robotic research. Traditionally, 6D object manipulation has been tackled via collision-based motion planning in the configuration space of a scene constructed by another perception system or manual annotations~\cite{Chitta:etal:2012, Diankov:Kuffner:2008}. Although this approach has been widely used in industry, the planning component requires complete (i.e., no missing objects) and fairly accurate scene understanding, which is challenging in a cluttered scene. There have been several attempts to increase the robustness of visual perception~\cite{Tremblay:etal:ARXIV2018, Chen:etal:IROS2019, Sui:etal:RAL2020}; however, object detection and pose estimation can still struggle with complex object piles due to close contacts and occlusions. To handle imperfect perception, recent work~\cite{Schwarz:etal:ICRA2017, Wada:etal:CVPR2020, Zeng:etal:ICRA2017} integrated perception with heuristics-based motion generation, which works with an uncertain estimate of object configurations. Although promising, this heuristics-based approach struggles to handle arbitrary object configurations and requires restricting the scene and target object that can deal with (e.g., requiring exhaustive distractor removal before picking the target). In this work, we use learning-based motion planning to maintain robustness even with imperfect perception.

\subsection{Learning robotic manipulation}

Recently, the use of deep learning for robotic manipulation has become prevalent with its progress in visual recognition with convolutional networks and observation-to-action policy learning~\cite{Levine:etal:JMLR2016, Zeng:etal:TRO2020, James:etal:ARXIV2021}. With convolutional networks, prior work~\cite{Levine:etal:IJRR2018, Pinto:Gupta:ICRA2016, Zeng:etal:ICRA2018} demonstrated robotic grasping from a single view without explicitly modeling object geometry. For optimizing the policy for a sequential motion, other work has applied deep reinforcement learning for discriminate (i.e., targeted) grasping~\cite{Devin:etal:CORL2018, Fang:etal:ICRA2018}, indiscriminate grasping~\cite{Kalashnikov:etal:CORL2018, Zeng:etal:IROS2018}, and retrieval~\cite{Kurenkov:etal:ARXIV2020}. Our work is along the line with the work on discriminative object manipulation~\cite{Devin:etal:CORL2018, Fang:etal:ICRA2018, Kurenkov:etal:ARXIV2020}, but instead focusing on how to manipulate objects after grasping, which has not been well explored in previous work. Moreover, we exploit object-level scene understanding in the learning-based manipulation model by feeding estimated poses along with a raw depth observation.

\section{Object-level Mapping}

To build a map of objects and find a target object in a scene using an RGB-D camera sequence, we use object-level mapping similar to prior work, MoreFusion~\cite{Wada:etal:CVPR2020}.  The mapping process consists of 1) single-view object detection with learning-based object detection; 2) occupancy-based volumetric reconstruction for object tracking and multi-view fusion, and 3) pose estimation using the volumetric reconstruction. To find a specific object, we query the object geometry from the map using the class ID of the object, which provides its mask and estimated pose.

\subsection{Object detection from a single view}

We use a state-of-the-art object detection model, Mask R-CNN~\cite{He:etal:ICCV2017}, which receives an RGB image and predicts object classes and masks. Using the detected mask, we extract the visible surface of an object from a depth image and accumulate it in the following multi-view volumetric fusion.

Like prior work, we train an object detection model for the YCB objects~\cite{Calli:etal:ICAR2015} using existing datasets~\cite{Wada:etal:CVPR2020, Xiang:etal:RSS2018}, which contains both real and synthetic images. Although this model is fairly accurate, it can generate false positives with a low threshold of the detection confidence, so we use a relatively high confidence threshold of 75\%, as per~\cite{Wada:etal:CVPR2020}. Although this high threshold favors false negatives in detecting objects, the multi-view mapping allows the model to run in different frames to find objects with a confident prediction.

\subsection{Volumetric reconstruction in multiple views}

We use occupancy volumetric reconstruction to track objects and accumulate their depth observation in multiple views. For efficient accumulation, we use an Octree-based occupancy volumetric fusion framework~\cite{Hornung:etal:AR2013}. Octree structure efficiently queries the associated voxel $v$ for a 3D query point in a new observation $z_t$, and the occupancy probability of the voxel $p(v|z_{1:t})$ at time $t$ is updated with a new observation $p(v|z_t)$ with Baysian update:
\begin{eqnarray}
  \mathcal{L}(v|z_{1:t+1}) = \mathcal{L}(v|z_{1:t}) + \mathcal{L}(v|z_t) \\
  \mathrm{where}\ \ \mathcal{L}(x) = \mathrm{logit}(p(x)) = \log(p(x) / (1 - p(x))). \nonumber
\end{eqnarray}
As we collect observations $z_t$ with a moving camera, we fuse them into an Octree-based volume, which filters sensor noises to build accurate reconstruction.

To associate the 3D point from an observation $z_t$ to the corresponding voxel $v^{o} \in V^o$ of an object $o$, we have two tracking mechanisms for a camera and objects. Camera tracking gives the transformation of an observation $z_t$ to the global coordinate system (i.e., map frame), and object tracking gives the corresponding octree map $V^o$ to which a new observation is accumulated. For camera tracking, we use forward kinematics of a robotic arm and the rigid transformation from the robot to the attached camera. For object tracking, we render the mapped objects in the live camera frame $M^o$. This rendered mask is compared with the detected masks $D_i$ by object detection with computing the intersection over the union (IoU) between them: $(M^o \land D_i) / (M^o \lor D_i)$. When the IoU is over a threshold of $0.4$, $D_i$ is recognized as another observation of the object $o$ and accumulated into its octree $V^o$ along with a observed depth, otherwise a new octree is initialized to map the new object.

\subsection{Pose estimation}
Using the volumetric reconstruction of objects, we estimate an object pose to replace it with a CAD model. This CAD model replacement gives full geometry of an object, whereas the volumetric reconstruction only gives partial geometry of the visible surface. We use a state-of-the-art pose estimation model in MoreFusion~\cite{Wada:etal:CVPR2020}, trained for the YCB objects~\cite{Calli:etal:ICAR2015}. Following \cite{Wada:etal:CVPR2020}, we exploit different views from a moving camera to acquire confident pose estimates. Given a set of estimated poses, we compute the point-to-point distance among the CAD models transformed with the poses. We replace the volumetric reconstruction with a CAD model when this process finds several matches.

\section{Learning Object Extraction}

To train the motion planning model, we use deep Q-learning~\cite{Mnih:etal:Nature2015}, an off-policy, model-free reinforcement learning algorithm. This algorithm learns a policy that maximizes the cumulative reward in an episode, using the collected episodes through exhaustive exploration in action space. We use the translations of objects as the negative reward (i.e., penalty) in this algorithm so that the model learns manipulation that minimizes destructive effects.

\subsection{Grasp point selection}

We select the centroid of the visible surface of a target object as the grasp point: $\vecp = [p_x, p_y, p_z]^\intercal$. The surface geometry and mask of a target object are extracted from the object-level map created in the preceding process. For the orientation, we compute a quaternion $\vecq = [q_x, q_y, q_z, q_w]^\intercal$ that gives the minimal transformation to align the axis of a suction cup $\vecv_g$ to the surface normal of the grasp point $\vecv_s$:
\begin{eqnarray}
  [q_x, q_y, q_z]^\intercal &=& \vecv_g \times \vecv_s \\
  q_w &=& \sqrt{\sum_i{\vecv_{g,i}^2} + \sum_i{\vecv_{s,i}^2}} + (\vecv_g^\intercal \cdot \vecv_s).
\end{eqnarray}

\subsection{Fusing raw and pose observations}

We take advantage of both raw and pose observations by feeding them as input to the model. The raw observation is formed as a heightmap generated from depth images, and the pose observation is extracted from the object-level map. These two observations provide different properties. The raw observation has few processes before being fed into the model, so it is less prone to estimation errors. In contrast, pose observation gives complete semantic and geometry (e.g., occluded parts of the objects), which is missing in the raw observation. The fusion of these two observations enables the model to achieve both high performance in object extraction (less destructive effects on a pile) and robustness to estimation errors (e.g., misdetection, pose-difference).

\subsection*{Raw observation}
For raw observation, we build a heightmap from depth images. The heightmap represents the heights of object surfaces on a planar workspace and gives information about the visible surface of objects. We build the heightmap centering the XY coordinates of the grasp point $p_x, p_y$ in the image coordinate of the heightmap for canonicalization regarding the grasp point. We use $0.004 m$ as the size that each pixel represents and 128 as the image height and width dimensions, giving $\pm0.256 m$ ($=0.004 \cdot 128 / 2$) XY bounds.

\subsection*{Pose observation}
As for the pose observation at training time, we extract the ground truth object poses and classes from the simulator that we use to train the manipulation model.  At test time, we extract object poses from the object-level map, which we build using an on-board camera on a robot.  We represent the pose observation with target flags (O,) with a binary vector, object classes (O, C) with one-hot vectors, and poses (O, 7) as a pair of position (3,) and quaternion (4,), where O is the number of detected objects and C is the number of classes.

Before we feed the pose observation into the model, we canonicalize it similarly to the heightmap by centering the grasp point in the pose coordinate. We subtract the XY coordinates of the grasp point $p_x$, $p_y$ from each pose of the object, which aligns the pose observation along to the raw observation (heightmap).

\subsection{Model design}\label{section:model_design}

The model represents a Q-function that predicts the discounted return given an observation $o_t$ at time $t$. Once trained, we evaluate this Q-function over a set of actions $a \in A$ to take the highest-valued one $\hat{a}$ (the next best action):
\begin{equation}
  \hat{a} = \argmax_a Q(a, o_t).
\end{equation}

\subsection*{Action}
We formulate the action as a relative 6D end-effector transformation discretizing each axis of translation and rotation.  We discretize the translation space in increments of $0.05$ m and discretize the Euler rotation space in increments of $22.5$ degrees. Each axis has either 0, positive or negative value, and their combinations give $3^6 = 729$ actions.

This relative end-effector action is taken $N$ times to construct a 6D arm trajectory. To give the information of previously taken actions, we feed the previous end-effector poses as input to the model along with the other visual observations (a heightmap; object poses).

\subsection*{Network architecture}
\begin{figure}[htbp]
  \centering
  \includegraphics[width=\linewidth]{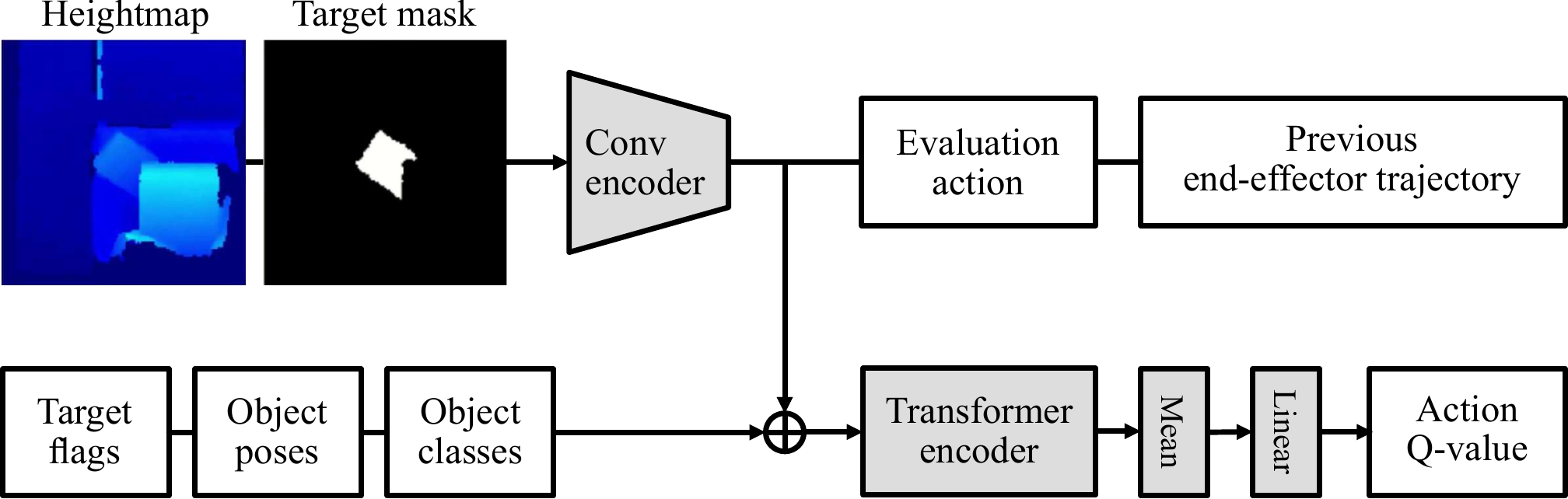}
  \caption{\textbf{Network architecture}, \small{which uses a heightmap and object poses to predict Q-value of 6D end-effector actions. We feed end-effector relative motions as evaluation actions, from which the best-scored action is selected as the next action.}}
  \label{figure:network}
  \vspace{1mm}\hrule\vspace{0mm}
\end{figure}

\figref{figure:network} shows the network architecture. The input heightmap is downsampled by a convolutional (Conv) encoder to a feature vector. This feature vector is concatenated with object poses, and processed by a transformer encoder~\cite{Vaswani:etal:NIPS2017}. With this transformer encoder, the model can handle an arbitrary number and order of object poses given as input. The output features from this encoder are averaged (similar to \cite{Baker:etal:ICLR2020}) to predict Q-value with a linear layer.

Along with the observations, the model receives two inputs: an evaluation action and a previous end-effector trajectory.  The evaluation action is fed as a relative translation and rotation. The previous end-effector trajectory represents the actions the model has taken with a list of 6D poses, allowing it to generate a consistent next action.

\subsection*{Reward}

The destructive effects on surrounding objects during manipulation can be classified as follows:
\begin{itemize}
  \item Falling, which happens when overlapping objects are mainly supported by a target object and they fall after the extraction of the target;
  \item Sliding, which happens when surrounding objects have to be displaced to create a space to extract a target.
\end{itemize}
Both of these two effects are undesirable. The falling effect can damage objects, and the sliding effect can expand the pile, which can make subsequent task continuation harder.

To cover both effects in the reward, we use the sum of the translations of non-target objects in a pile as the metric. When an object falls a large distance, not only does the translation of the object itself become significant, but it can also hit other objects causing chain effects.  By using the ``sum" of translations, we can encourage the model to minimize the number of objects affected as well as the translations of individual objects.

\subsection{Training the model}

We compute the reward $r_t$ at each time step $t$, and train a deep Q-network to maintain the Bellman equation:
\begin{eqnarray}
  q_{t,a} = Q(o_t, a) \\
  \hat{q}_{t,a} = r_t + \gamma \max_a\tilde{Q}(o_{t+1}) \\
  \mathcal{L} = |\hat{q}_{t,a} - q_{t,a}|,
\end{eqnarray}
where $\gamma$ is the time discount of the reward, Q is the live network updated every training step, and $\tilde{Q}$ is the target network, a copy of the live network updated less frequently.

We train the model in a physics simulation environment since, in the real world, it is difficult to track the motions of objects to compute the reward. It would be also time-consuming and challenging to build various configurations of objects for each trial of a robot learning in the real world while maintaining the safety of the robot and objects.

To let a robot experience various configurations of objects, we procedurally generate object piles by simulating feasible pile configurations. We define a 3D bounding box where object models can be spawned. Each step randomly selects a model and its pose, and we apply physics simulation until the spawned object stops moving. This process produces diverse object configurations.

\section{Experiments}
We evaluate our method by assigning a robot to grasp and extract a target object from object piles in both simulation and the real world.  For training the learning model in simulation, we use the YCB models provided by~\cite{Calli:etal:ICAR2015, Xiang:etal:RSS2018}.

\begin{figure*}[t]
  \centering
  \includegraphics[width=\linewidth]{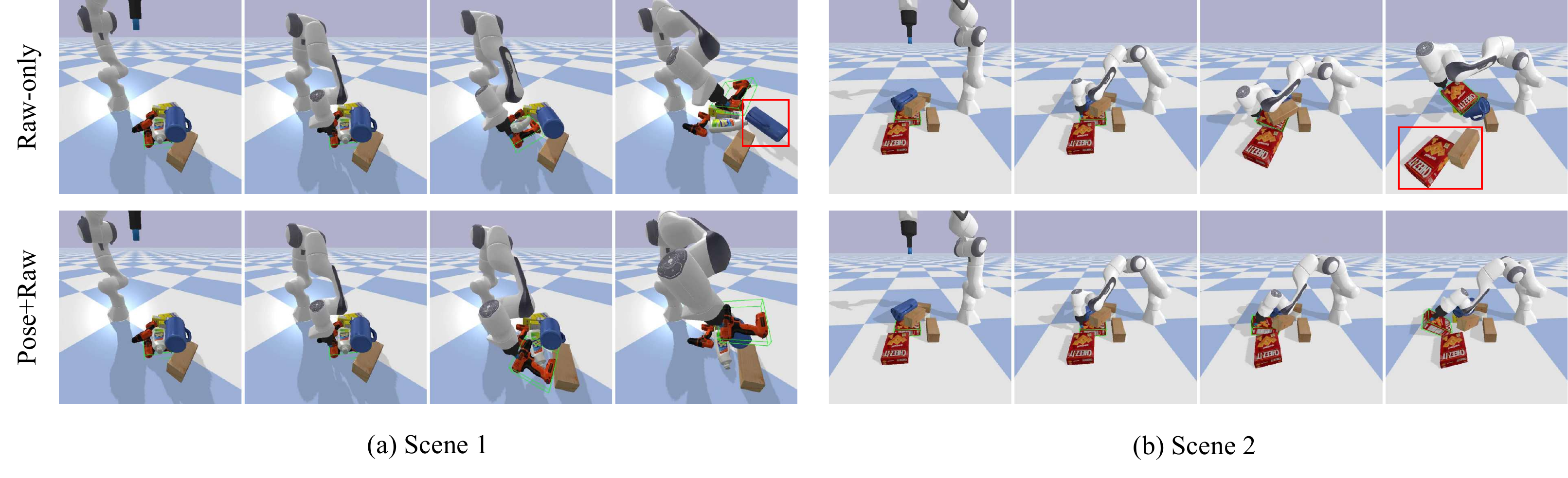}
  \caption{\textbf{Qualitative results of the model ablation}, \small{in which we compare \textit{Raw-only}, which uses only a heightmap observation, and our full model \textit{Pose+Raw}, which uses object poses along with a heightmap.}}
  \label{figure:qualitative_comparison}
  \vspace{1mm}\hrule\vspace{-0mm}
\end{figure*}

\subsection{Training detail}
We implement the learning model with PyTorch~\cite{Paszke:etal:ANIPS2019} and the simulation environment with PyBullet~\cite{Erwin:Yunfei:Misc2016}. We run a single process to update the model parameters using action-state paired data. This data is collected by multiple processes that asynchronously run the learned model to act in different environments for exploration. For this asynchronous data collection and training, we use an open-source framework~\cite{James:YARR}.

As for the training hyperparameters, we use a batch size of 128, and Adam~\cite{Kingma:Ba:ICLR2015} as the optimizer with a learning rate of 0.001.  From the start of training, we use epsilon greedy exploration to collect data until 5000 iterations. We use replay ratio 16 (number of updates per data collection), and synchronize the model parameters every 100 iterations to allow the model in the exploration processes to collect episodes with the new parameters.

\subsection{Evaluation metrics in simulation}
To evaluate the performance of the model, we define metrics that represent how safely a robot extracted objects. As we discussed in \secref{section:model_design}, this metric should represent the destructive effects on non-target objects caused by falling and sliding, and for that, we use the followings:
\begin{itemize}
  \item Sum of translations, which evaluates both falling and sliding effects;
  \item Sum of max velocities, which primarily evaluates falling effects, as a larger distance fall gives higher velocity.
\end{itemize}

\subsection{Baseline comparison}

\subsubsection{Naive motion} As the simplest motion, we use joint linear interpolation from a grasp pose to a reset pose where the end-effector is located in free space and the suction cup faces down.  When a target object is overlapped by distractor objects, this extraction motion introduces many collisions and causes falling and sliding effects on the distractors, and shows the lowest score in \tabref{table:baseline_comparison}.

\subsubsection{Heuristic motion} As a simple heuristic for extracting objects from a pile, we use a motion that extracts a grasped object with a straight motion towards the +Z direction of the world coordinate (the opposite direction of the gravity). This motion gives better results than the naive motion (\tabref{table:baseline_comparison}).

\subsubsection{Collision-based motion planning} As another baseline, we use a collision-based motion planning, RRT-Connect~\cite{Kuffner:etal:ICRA2000}, which is a widely used motion-planning algorithm with estimated object poses. Although this motion planning can give non-destructive extraction motion when it finds a collision-free path, it struggles to find \textit{relatively} safe trajectories when a complete collision-free path does not exist. In this case, the motion planner can end up giving the naive motion, which causes significant movement of surrounding objects. \tabref{table:baseline_comparison} shows that RRT-Connect gives comparable results as the heuristic motions (better in translation, worse in velocity), but our learned model (SafePicking) gives a much better result.

\begin{table}[htbp]
  \centering
  \caption{\textbf{Baseline comparison}\small{, in which we compare the motion of our learned model with baseline methods using the safety metric, in 600 unseen pile configurations in simulation.}}
  \label{table:baseline_comparison}
  \begin{tabular}{lcc|cc}
    & & & \multicolumn{2}{c}{\textbf{Safety metric}} \\
    \textbf{Method}              & \textbf{Input}        &
    \textbf{Noise} & \textbf{translation$\downarrow$} & \textbf{velocity$\downarrow$} \\ \Xhline{2\arrayrulewidth}
  Naive       & - & \multirow{4}{*}{no} & 0.701  & 1.919 \\
  Heuristic   & - &  & 0.578  & 1.624 \\
  RRT-Connect & pose &  & 0.520  & 1.643 \\
  SafePicking & pose, heightmap &  & \textbf{0.465}  & \textbf{1.419} \\ \hline
  RRT-Connect & pose & \multirow{2}{*}{yes} & 0.532  & 1.645 \\
  SafePicking & pose, heightmap &  & \textbf{0.465}  & \textbf{1.433} \\
  \end{tabular}
\end{table}

\subsection{Model ablation}
We evaluate 3 model variants with different inputs to compare the effect of different observations given to the learned model (\tabref{table:model_comparison}). To evaluate the robustness, we add noises to the pose observation (misdetection, pose-difference) based on the visibility of objects. \textit{Raw-only} receives as input only the heightmap generated from a depth image.  \textit{Pose-only} uses object poses as input, which gives more complete information about objects, and gives better results than Raw-only even when noises are introduced in the input object poses. The full model, \textit{Pose+Raw}, receives both heightmap and object pose as input, and compensates the noises in object poses using the information about the object's visible surface from the heightmap. The results show that this fusion of object pose and heightmap observations works best in the presence of pose noises. As one would expect, when given perfect pose information (no pose noises), there is no benefit to having the additional heightmap information; however, in reality, estimated object poses will always have errors. \figref{figure:qualitative_comparison} shows qualitative results.

\begin{table}[htbp]
  \centering
  \caption{\textbf{Model ablation}\small{, in which we compare the variants of
  the learned model with/without adding noises to the pose observation, and
  test in 600 unseen pile configurations in simulation.
  }}
  \label{table:model_comparison}
  \begin{tabular}{lcc|cc}
    & & & \multicolumn{2}{c}{\textbf{Safety metric}} \\
  \textbf{Variant} & \textbf{Input} & \textbf{Noise} & \textbf{translation$\downarrow$} & \textbf{velocity$\downarrow$} \\ \Xhline{2\arrayrulewidth}
    Raw-only  & heightmap & \multirow{3}{*}{no} & 0.507 & 1.491 \\
    Pose-only & pose & & 0.477 & 1.430 \\
    Pose+Raw & pose, heightmap & & \textbf{0.465} & \textbf{1.419} \\  \hline
    Pose-only & pose & \multirow{2}{*}{yes} & 0.487 & 1.449 \\
    Pose+Raw & pose, heightmap & & \textbf{0.465} & \textbf{1.433} \\
  \end{tabular}
\end{table}

\subsection{Real-world evaluation}

We evaluate our system in the real world on a Franka Emika Panda robot,
integrating using the Robotic Operation System (ROS)
framework~\cite{Quigley:etal:ICRA2009}.  To capture the short-range
depth ($\sim$0.1m), we used Realsense
D435~\cite{Keselman:etal:CVPRW2017} as the RGB-D camera mounted on the
wrist of the robotic arm.

\subsubsection{Metric}
\begin{figure}[htbp]
  \centering
  \includegraphics[width=\linewidth]{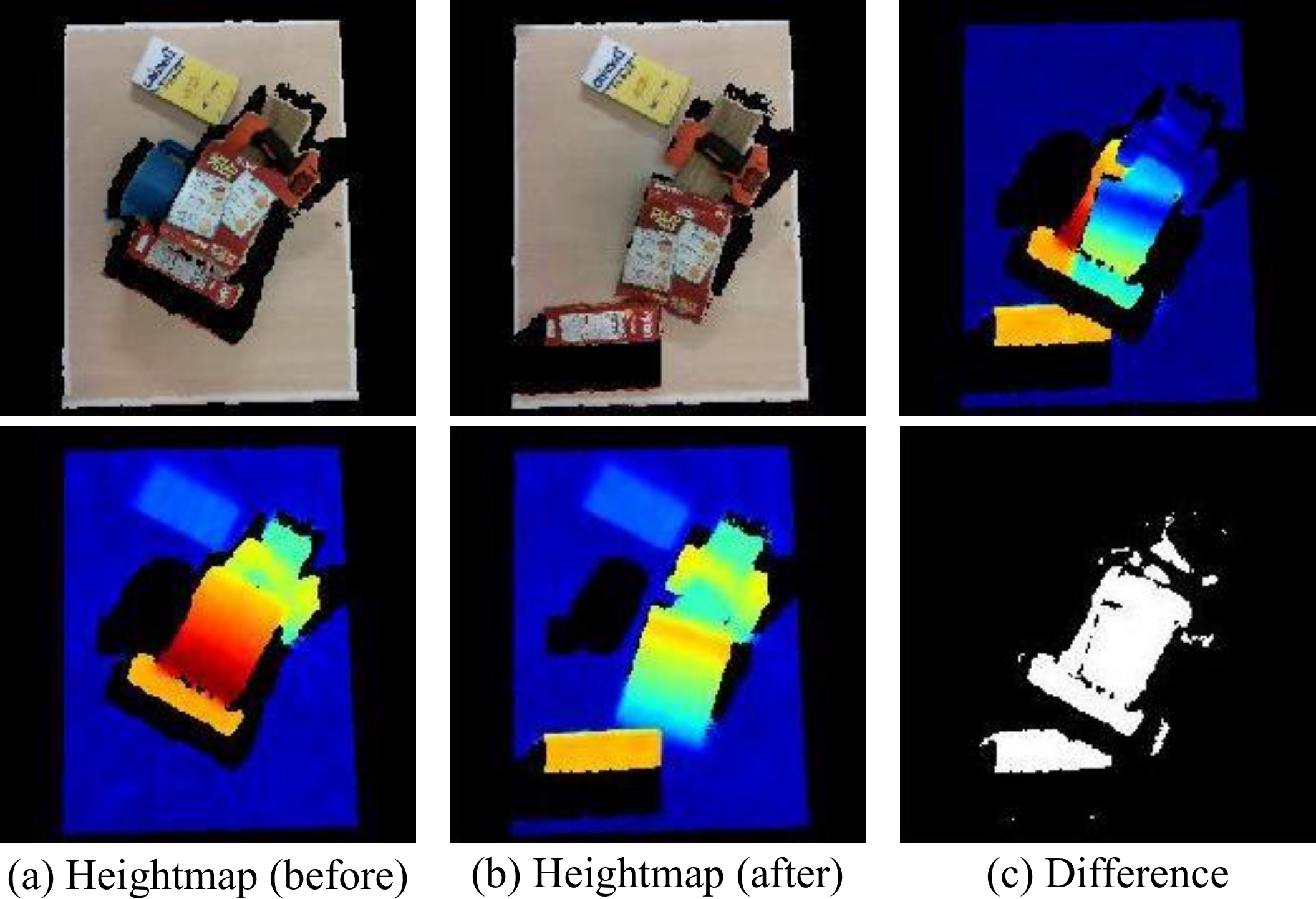}
  \caption{\textbf{Heightmap-based evaluation metric}\footnotesize{ that compares the heightmaps of a pile built before/after manipulation. The region of a target object (blue pitcher in this case) is excluded from the comparison and is filled with black color in the heightmap visualization.}}
  \label{figure:quantitative_real}
  \vspace{1mm}\hrule
\end{figure}

In the real world, the safety metrics used in the simulation (the translation and velocity of all objects) are challenging to acquire. Therefore, we use the heightmap difference between before and after a task as the metric (\figref{figure:quantitative_real}). The robot scans a pile and builds two heightmaps (\figref{figure:quantitative_real}a, b), in which the region of a target object is excluded (since the target object is intentionally moved). These heightmaps are compared (\figref{figure:quantitative_real}c) to compute the volume of the difference and the size of the mask at a threshold of $0.01m$.

\subsubsection{Results}
\begin{figure}[htbp]
  \centering
  \includegraphics[width=0.98\linewidth]{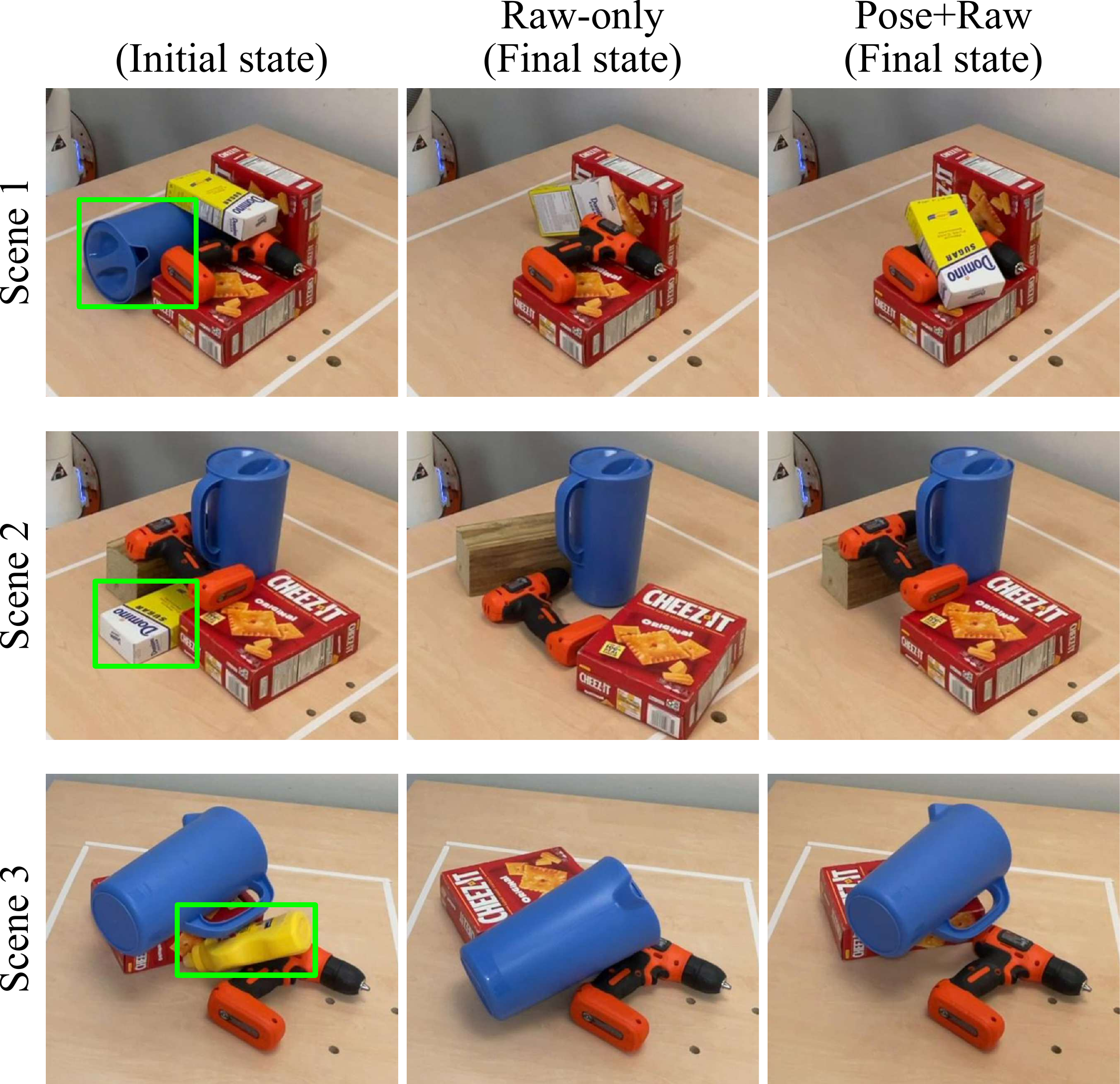}
  \caption{\textbf{Pile configurations}\footnotesize{ out of 20 used in \tabref{table:model_comparison_real}, where we test the two learned models. We measure the performance with the heightmap-based safety metric. Consistently to the results in the simulation, the model generates better motions when given a fused observation of object poses and a heightmap (Pose+Raw).}}
  \label{figure:piles_real}
  \vspace{1mm}\hrule\vspace{-2mm}
\end{figure}

Using the heightmap-based metric, we evaluate the variants of the learned models in the real world.  As for the task setting, we build similar object configurations as the simulation, where objects are closely located in a pile and a target object is partially occluded by other objects, and the robot is tasked to extract the target object with a single grasp. \tabref{table:model_comparison_real} shows the comparison of the two learned models (Raw-only and Pose+Raw) in 20 configurations. Some of their qualitative results are shown in \figref{figure:piles_real}. These comparisons show consistent results as the ablation study in the simulation, showing that object pose information enables the model (Pose+Raw) to generate efficient and safe motions compared to the one with only raw observations (Raw-only).

\begin{table}[htbp]
  \centering
  \caption{\textbf{Real-world model comparison}\footnotesize{, in which we evaluate learned models comparing the heightmaps before/after each task. The models are tested in the same 20 configurations. }}
  \label{table:model_comparison_real}
  \begin{tabular}{c|cc}
    & \multicolumn{2}{c}{\textbf{Safety metric}} \\
    \textbf{Variant} & \textbf{Diff mask [\%]$\downarrow$} & \textbf{Diff
    volume [litter]$\downarrow$} \\ \hline
    Raw-only & 7.1 & 3.2 \\
    Pose+Raw & \textbf{4.4} & \textbf{2.1} \\
  \end{tabular}
\end{table}

\section{Conclusions}

We have shown a 6D manipulation system to efficiently and safely extract target objects from a pile with a trajectory that minimizes destructive effects on the surrounding objects.  Our system integrates object-level mapping and learning-based motion planning that uses raw observation from an RGB-D camera and estimated object poses from the object-level map. By fusing the raw and pose observations, the motion model maintains both capabilities of safely extracting objects and robustness to the errors in pose estimation.

In this work, we focused on post-grasp 6D manipulation of household objects; however, it would be interesting to extend this to long-term motion optimization with grasping and placement, and more general objects including soft and high-friction objects. Moreover, we believe there are still various possibilities in integrating object-level scene understanding with learning-based manipulation. This work has mainly exploited the completeness of pose estimates; however, object pose information can also be used in different ways such as for object placement in a specific pose.

\section*{Acknowledgements}
Research presented in this paper has been supported by Dyson Technology Ltd.

\bibliographystyle{plain}
\bibliography{robotvisiontex/robotvision,root}

\end{document}